\documentstyle[aclap]{article}

\newcommand{\ovis}{OVIS}

\title{\vspace{-0.5in}Grammatical analysis in the \ovis\ spoken-dialogue 
system\thanks{\ \ To appear in the proceedings of the
ACL/EACL Workshop on Spoken Dialog Systems, Madrid, Spain - July 11-12, 1997.}}
\author{%
\bf Mark-Jan Nederhof \\
\bf Gosse Bouma  \\
\bf Rob Koeling \\
\bf Gertjan van Noord \\ \\
Faculty of Arts, Humanities Computing \& BCN\\
University of Groningen \\
P.O.\ Box 716, NL-9700 AS Groningen, The Netherlands \\
\{markjan,gosse,koeling,vannoord\}@let.rug.nl}

\begin{document}

\maketitle
\vspace{-0.5in}
\begin{abstract}

We argue that grammatical processing is a viable
alternative to concept spotting for processing spoken input
in a practical dialogue system.
We discuss the structure of the grammar, the properties of the
parser, and a method for achieving robustness.  We discuss test
results suggesting that grammatical processing allows fast and
accurate processing of spoken input.

\end{abstract}

\section{Introduction}
\label{intro}

The NWO Priority Programme {\em Language and Speech Technology\/} is a
research programme aiming at the development of spoken language
information systems. Its immediate goal is to develop a demonstrator
of a public transport information system, which operates over ordinary
telephone lines.  This demonstrator is
called \ovis, Openbaar Vervoer Informatie Systeem ({\em Public
Transport Information System\/}). The language of the system is Dutch.

At present, a prototype is in operation, which is a version of a 
German system developed by Philips
Dialogue Systems in Aachen \cite{AU95}, adapted to Dutch.

This German system processes spoken input using ``{\em concept
spotting\/}'', which means that the smallest information-carrying
units in the input are extracted,
such as names of train stations and expressions of
time, and these are translated more or less individually into updates of the
internal database representing the dialogue state.
The words between the concepts thus perceived are ignored.

The use of concept spotting is common
in spoken-language information systems \cite{WA89,JA91,AU95,AL96}.
Arguments in favour of this kind of shallow parsing is that it is
relatively easy to develop the NLP component, since larger sentence constructs
do not have to be taken into account, and that the robustness of the
parser is enhanced, since sources of ungrammaticality occurring between
concepts are skipped and therefore do not hinder the translation of the
utterance to updates.

The prototype presently under construction departs from the use of
concept spotting. The grammar for \ovis\ describes {\em grammatical\/}
user utterances, i.e.\ whole sentences are described. Yet,
as part of this it also
describes phrases such as expressions of time and prepositional
phrases involving e.g.\ train stations, in other words, the former
concepts. By an appropriate parsing algorithm one thus combines the 
robustness that can be achieved using concept spotting with 
the flexibility of a sophisticated language model.

The main objective of this paper is to show that our grammatical approach 
is feasible in terms of accuracy
and computational resources, and thus is a viable alternative 
to pure concept spotting. 

Although the added benefit of grammatical analysis over concept
spotting is not clear for our relatively simple
application, the grammatical approach may become essential
as soon as the application is extended in such a way that
more complicated grammatical constructions need to be recognized.
In that case, simple concept spotting may not be able to correctly
process all constructions, whereas the capabilities of the
grammatical approach extend much further.

Whereas some (e.g.\ \cite{MO89a}) argue that grammatical 
analysis may improve recognition accuracy, 
our current experiments have as yet not been able to reveal a 
clear advantage in this respect.

As the basis for our implementation we have chosen definite-clause
grammars (DCGs) \cite{PE80}, 
a flexible formalism which is related to various kinds of
common linguistics description, and which allows application of 
various parsing algorithms. DCGs
can be translated directly into Prolog, for which interpreters and 
compilers exist that are
fast enough to handle real-time processing of spoken input.
The grammar for OVIS is in turn written in a way to allow
an easy translation to pure DCGs.\footnote{DCGs are called {\em pure\/}
if they do not contain any calls to external Prolog predicates.}

The structure of this paper is as follows. In Section~\ref{grammar}
we describe the grammar for OVIS, and in Section~\ref{dialogue}
we describe the output of the NLP module. 
The robust parsing algorithm is described in Section~\ref{parser}.
Section~\ref{eval} reports test results, showing that
grammatical analysis allows fast and accurate processing of 
spoken input.

\section{A computational grammar for Dutch}
\label{grammar}

In developing the \ovis\ grammar we have tried to combine the short-term goal
of developing a grammar which meets the requirements imposed by the application
(i.e.\ robust processing of the output of the speech recognizer, 
extensive coverage of 
locative phrases and temporal expressions, 
and the construction of fine-grained semantic 
representations) with the long-term goal of developing a general, 
computational, grammar which covers all the major constructions of Dutch. 

The grammar currently covers the majority of verbal
subcategorization types (intransitives, transitives, verbs selecting a
{\sc pp}, and modal and auxiliary verbs), {\sc np}-syntax
(including pre- and postnominal modification, with the exception of
relative clauses), {\sc pp}-syntax, the distribution of 
{\sc vp}-modifiers, various clausal types (declaratives, yes/no and 
{\sc wh}-questions, and subordinate clauses), all temporal expressions and
locative phrases relevant to the domain, and various typical 
spoken-language constructs. 
Due to restrictions imposed by the speech recognizer,
the lexicon is relatively small (2000 word forms, most of which are 
names of stations and cities).

{}From a linguistic perspective, the \ovis-grammar can be
characterized as a constraint-based grammar, which makes heavy use of (multiple)
inheritance. As the grammar assumes quite complex lexical signs, inheritance is
absolutely essential for organizing the lexicon succinctly.
However, we not only use inheritance at the level of the lexicon (which is a
well-known approach to computational lexica), but have also structured the 
rule-component using inheritance. 

An important restriction imposed by the 
gram\-mar\discretionary{-}{}{-}par\-ser 
interface is that rules must specify the category 
of their mothers and
daughters. That is, each rule must specify the type of sign
of its mother and daughters. A consequence of this requirement is that
general rule-schemata, as used in Categorial Grammar and {\sc hpsg}
cannot be used directly in the \ovis\ grammar. A rule which specifies that
a head daughter may combine with a complement daughter, if this
complement unifies with the first element on {\sc subcat} of the head
(i.e.\ a version of the categorial rule for functor-argument
application) cannot be implemented directly, as it leaves the
categories of the daughters and mother unspecified. Nevertheless,
capturing generalizations of this type does seem desirable.  

We have therefore adopted an architecture for grammar rules similar to 
that of {\sc hpsg} \cite{PO94}, in which individual rules are classified in 
various {\em structures}, which are in turn defined in terms of general 
{\em principles}.  For instance, the grammar currently contains several
head-complement rules (which allow a verb, preposition, or determiner to 
combine with one or more complements). These rules need only specify 
category-information 
and the relative order of head and complement(s). 
All other information associated 
with the rule (concerning the matching of head-features, the instantiation of 
features used to code long-distance dependencies, 
and the semantic effect of the 
rule) follows from the fact that the rules are instances of the 
class {\em head-complement structure}. This class itself is defined in terms of
general principles, such as the {\em head-feature}, {\em valence}, {\em
filler} and {\em semantics} principle. Other rules are defined in terms of the
classes {\em head-adjunct} and {\em head-filler structure\/}, which in turn
inherit from (a subset of) the general principles mentioned above. Thus,
even though the grammar contains a relatively large number of rules (compared to
lexicalist frameworks such as {\sc hpsg} and {\sc cg}), the redundancy in these
rules is minimal.

The resulting grammar has the interesting property that it combines the strong
tendency towards lexicalism and positing general combinatoric rule schemata
present in frameworks such as {\sc hpsg} with relatively specific 
grammar rules to facilitate efficient processing. 

\section{Interaction with the dialogue manager}
\label{dialogue}

The semantic component of the grammar produces (simplified)
Quasi-Logical Forms \cite{AL92}. These are linguistically motivated,
domain-independent representations of the meaning of utterances.  

{\sc qlf}s allow considerable underspecification. 
This is convenient in this application because most ambiguities
that arise, such as ambiguities of scope, do not need to be resolved.
These {\sc qlf}s are
translated into domain-specific ``updates'' to be passed on to the
dialogue manager ({\sc dm}) for
further processing. The {\sc dm} keeps track of the information
provided by the user by maintaining an {\em information state\/} or {\em
form}. This form is a hierarchical structure, with
slots and values for the origin and destination of a connection, for
the time at which the user wants to arrive or depart, etc. The
distinction between slots and values can be regarded as a special case
of ground and focus distinction \cite{vallduvi}.
Updates specify the ground and focus of the user utterances. For
example, the utterance 
{\it ``No, I don't want to travel to
Leiden but to Abcoude!''} 
yields the following update:

\begin{verbatim}
userwants.travel.destination.
                  ([# place.town.leiden];
                   [! place.town.abcoude])
\end{verbatim}

One important property of this representation is that it
allows encoding of speech-act information. The ``\verb|#|'' in the update
means that the information between the square brackets (representing
the focus of the user-utterance) must be retracted, while the ``\verb|!|''
denotes the corrected information.

\section{Robust parsing}
\label{parser}

The input to the NLP module consists of word-graphs produced by the 
speech recognizer. A word-graph is a compact representation for all
lists of words that the speech recognizer hypothesizes for a spoken 
utterance. The nodes of the graph represent points in time, and
an edge between two nodes represents a word that may have been uttered 
between the corresponding points in time. Each edge is associated with
an {\em acoustic score} representing a measure of confidence that the word 
perceived there is the word that was actually uttered. These scores
are negative logarithms of probabilities and therefore require
addition as opposed to multiplication when two scores are combined.

At an early stage, the word-graph is optimized to eliminate the
{\em epsilon transitions}. Such transitions represent periods of time when the
speech recognizer hypothesizes that no words are uttered. After 
this optimization, the word-graph contains exactly one start node and
one or more final nodes, associated with a score,
representing a measure of confidence that the utterance ends at that point.

In the ideal case, the parser will find one or more paths in
a given word-graph that can be assigned an analysis according to the
grammar, such that the paths cover the complete time span of the
utterance, i.e.\ the paths lead from the start node to a final
node. Each analysis gives rise to an update of the dialogue
state. From that set of updates, one is
then passed on to the dialogue manager.

However, often no such paths can be found in the word-graph,
due to:
\begin{itemize}
\item errors made by the speech recognizer,
\item linguistic constructions not covered in the grammar, and
\item irregularities in the spoken utterance.
\end{itemize}

Our solution is to allow recognition of paths in the word-graph
that do not necessarily span the complete utterance. Each path should 
be an instance of some major category from the grammar, such as
{\sc s}, {\sc np}, {\sc pp}, etc. 
In our application, this often comes down to categories such as
``temporal expression'' and ``locative phrases''.
Such paths will be called {\em maximal projections}.
A list of maximal projections that do not pair-wise overlap and 
that lie on a single path from the start node
to a final node in the word-graph represents a reading of the utterance.
The transitions between the maximal projections will be called 
{\em skips}.

The optimal such list is computed, according to criteria to be 
discussed below.
The categories of the
maximal projections in the list are then combined and 
the update for the complete utterance is computed.
This last phase contains, among other things, some domain-specific linguistic 
knowledge dealing with expressions that may be ungrammatical in other 
domains; e.g.\ the utterance {\it ``Amsterdam Rotterdam''}
does not exemplify a general grammatical
construction of Dutch, but in the particular domain of OVIS such an
utterance occurs frequently, with the meaning
{\it ``departure from Amsterdam and arrival in Rotterdam''}.

We will now describe the robust parsing module in more detail.
The first phase that is needed is the application of a
parsing algorithm which is such that:
\begin{enumerate}
\item grammaticality is investigated for all paths, not only for the
complete paths from the first to a final node in the word-graph, and
\item grammaticality of those paths is investigated for each category
from a fixed set.
\end{enumerate}
Almost any parsing technique, such as left-corner parsing, LR parsing, etc.,
can be adapted so that the first constraint above is satisfied; the
second constraint is achieved by structuring the grammar such that
the top category directly generates a number of grammatical categories.

The second phase is the selection of the optimal list of maximal projections 
lying
on a single path from the start node to a final node. 
At each node
we visit, we compute a partial score consisting of a tuple $(S, P, A)$,
where $S$ is the number of transitions on the path not part of a 
maximal projection (the skips), 
$P$ is the number of maximal projections, 
$A$ is the sum of the acoustic scores of all the transitions on the path,
including those internal in maximal projections.
We define the relation $\prec$ on triples such that
$(S_1, P_1, A_1)\prec(S_2, P_2, A_2)$ if and only if:
\begin{itemize}
\item $S_1 < S_2$, or
\item $S_1 = S_2$ and $P_1 < P_2$, or
\item $S_1 = S_2$ and $P_1 = P_2$ and $A_1 < A_2$.
\end{itemize}
In words, for determining which triple has minimal score (i.e.\ is optimal),
the number of skips has strictly the highest importance, then the number of
projections, and then the acoustic scores. 

Our branch-and-bound algorithm maintains a priority queue, which 
contains pairs of the form $(N, (S,P,A))$, 
consisting of a node $N$ and a triple $(S,P,A)$ found at the node,
or pairs of the form $(\widehat{N}, (S,P,A))$, with the same meaning
except that $N$ is now a final node of which the acoustic score is 
incorporated into $A$.
Popping an element from the queue yields a pair of which the
second element is an optimal triple with regard to the relation $\prec$
fined above.
Initially, the queue contains just $(N_0, (0,0,0))$, where $N_0$ is 
the start node, and possibly $(\widehat{N_0}, (0,0,A))$, if 
$N_0$ is also a final state with acoustic score $A$.

A node $N$ is marked as {\it seen\/} when a triple has been encountered
at $N$ that must be optimal with respect to all paths leading to
$N$ from the start node.

The following is repeated until a final node is found with an optimal
triple:
\begin{enumerate}
\item Pop an optimal element from the queue.
\item If it is of the form $(\widehat{N}, (S,P,A))$ then return the path
leading to that triple at that node,
and halt.
\item Otherwise, let that element be $(N, (S,P,A))$.
\item If $N$ was already marked as {\it seen\/} then abort this
iteration and return to step 1.
\item Mark $N$ as {\it seen}.
\item For each maximal projection from $N$ to $M$ with acoustic score $A'$,
enqueue $(M, (S,P+1,A+A'))$. If $M$ is a final node with acoustic score
$A''$, then furthermore enqueue $(\widehat{M}, (S,P+1,A+A'+A''))$.
\item For each transition from $N$ to $M$ with acoustic score $A'$,
enqueue $(M, (S+1,P,A+A'))$. If $M$ is a final node with acoustic score
$A''$, then furthermore enqueue $(\widehat{M}, (S+1,P,A+A'+A''))$.
\end{enumerate}

Besides $S$, $P$, and $A$, 
other factors can be taken into account as well,
such as the {\em semantic\/} score, which is obtained by comparing the
updates corresponding to maximal projections with the meaning of the
question generated by the system prior to the user utterance.

We are also experimenting with the bigram score.
Bigrams attach a measure of likelihood to
the occurrence of a word given a preceding word.

Note that when bigrams are used, simply labelling nodes in the graph
as {\it seen\/} is not a valid method to prevent recomputation of
subpaths. The required adaptation to the basic
branch-and-bound algorithm is not discussed here.

Also, in the actual implementation the $X$ best readings are produced, 
instead of a single best reading. 
This requires a generalization of the above procedure
so that instead of using the label {\it ``seen''},
we attach labels {\it ``seen i times''} to each node, where $0\leq i \leq X$.

\section{Evaluation}
\label{eval}

This section evaluates the NLP component with respect to efficiency
and accuracy. 

\subsection{Test set}

We present a number of results to indicate how well the NLP component
currently performs. We used a corpus of more than 20K word-graphs,
output of a preliminary version of the speech recognizer, and typical
of the intended application. The first 3800 word-graphs of this set
are semantically annotated. This set is used in the experiments below.
Some characteristics of this test set are given in Table~\ref{tat}. As can
be seen from this table, this test set is considerably easier than the
rest of this set. For this reason, we also present results (where
applicable) for a set of 5000 arbitrarily selected word-graphs.  At
the time of the experiment, no further annotated corpus material was
available to us. 

\begin{table}[t]
\begin{center}
\begin{tabular}{|r|r|r|r|r|r|}\hline
     & graphs & transitions & words & t/w  & w/g\\\hline
 test &  5000  &    54687    & 16020 & 3.4 & 3.2\\
 test&   3800 &     36074    & 13312 & 2.7 & 3.5\\
total & 21288 &    242010    & 70872 & 3.4 & 3.3\\\hline
\end{tabular}
\end{center}
\caption{\label{tat}
This table lists the number of transitions, the number of words of the
actual utterances, the average number of transitions per word, and the
average number of words per utterances. }
\end{table}

\subsection{Efficiency}

We report on two different experiments.  In the first experiment, the
parser is given the utterance as it was actually spoken (to simulate a
situation in which speech recognition is perfect).  In the second
experiment, the parser takes the full word-graph as its input. The
results are then passed on to the robustness component. We report on
a version of the robustness component which incorporates
bigram-scores (other versions are substantially faster). 

All experiments were performed on a HP-UX 9000/780 machine with more
than enough core memory.  Timings measure CPU-time and should be
independent of the load on the machine. The timings include all phases
of the NLP component (including lexical lookup, syntactic and semantic
analysis, robustness, and the compilation of semantic representations
into updates). The parser is a head-corner parser implemented (in
SICStus Prolog) with selective memoization and goal-weakening as
described in \cite{NO97}.  Table~\ref{ovista} summarizes the results
of these two experiments.

\begin{table*}[tb]
\begin{center}
\begin{tabular}{|l|l|r|r|r|r|}\hline
& mode                &  total msec & msec/sent & max msec & max kbytes\\\hline
{\em 3800 graphs:} &
 {\em user utterance} &      125290 &        32 &      330 &         86\\
&{\em word-graph}     &      303550 &        80 &     8910 &       1461\\\hline
{\em 5000 graphs:} &
 {\em user utterance} &      152940 &        30 &      630 &        192\\
&{\em word-graph}     &      477920 &        95 &    10980 &       4786\\\hline
\end{tabular}
\end{center}
\begin{center}
\begin{tabular}{|l|r|r|r|r|r|r|}\hline
&  100 &   200  &   500 &  1000 & 2000  & 5000 \\\hline
{\em 3800 graphs:} &
80.6  &  92.4  & 98.2  & 99.5  & 99.9  & 99.9 \\\hline
{\em 5000 graphs:} &
81.3  &  91.2  & 96.9  & 98.7  & 99.5  & 99.9 \\\hline
\end{tabular}
\end{center}
\caption{\label{ovista}
In the first table we list respectively the total
number of milliseconds CPU-time required for all 3800 word-graphs, the
average number of milliseconds per word-graph, and the maximum number
of milliseconds for a word-graph.  The final column lists the maximum
space requirements (per word-graph, in Kbytes). For word-graphs the
average CPU-times are actually quite misleading because CPU-times
vary enormously for different word-graphs.  For this reason, we present
in the second table the proportion of word-graphs that can be
treated by the NLP component within a given amount of CPU-time (in
milliseconds).}
\end{table*}

{}From the experiments we can conclude that almost all input word-graphs
can be treated fast enough for practical applications.  In fact, we
have found that the few word-graphs which cannot be treated
efficiently almost exclusively represent cases where speech
recognition completely fails and no useful combinations of edges can
be found in the word-graph.  As a result, ignoring these few cases
does not seem to result in a degradation of practical system
performance.

\subsection{Accuracy}

In order to evaluate the accuracy of the NLP component, we used the
same test set of 3800 word-graphs. For each of these graphs we know
the corresponding actual utterances and the update as assigned by the
annotators.  We report on word and sentence accuracy, which is an
indication of how well we are able to choose the best path from the
given word-graph, and on concept accuracy, which indicates how often
the analyses are correct.

The string comparison on which sentence accuracy and word accuracy are
based is defined by the minimal number of substitutions, deletions and
insertions that is required to turn the first string into the second
(Levenshtein distance).  The string that is being compared
with the actual utterance is defined as the best path through the
word-graph, given the best-first search procedure defined in the
previous section.  Word accuracy is defined as
$ 1 -  \frac{d}{n} $ where $n$ is the length of the actual utterance
and $d$ is the distance as defined above.

In order to characterize the test sets somewhat further,
Table~\ref{tabb} lists the word and sentence accuracy both of the best
path through the word-graph (using acoustic scores only), the best
possible path through the word-graph, and a combination of the
acoustic score and a bigram language model. The first two of these can
be seen as natural upper and lower boundaries. 

\begin{table*}[tb]
\begin{center}
\begin{tabular}{|l|c|r|r|}\hline
& method               &        WA &      SA\\\hline
{\em 3800 graphs:} &
Acoustic             &      78.9 &    60.6\\
& Possible             &      92.6 &    82.7\\
& Acoustic + Bigram    &      86.3 &    74.3\\\hline
{\em 5000 graphs:} &
Acoustic             &      72.7 &    57.6\\
& Possible             &      89.8 &    81.7\\
& Acoustic + Bigram    &      82.3 &    74.0\\\hline
\end{tabular}
\end{center}
\caption{\label{tabb}Word accuracy and sentence accuracy based on
  acoustic score only (Acoustic); using the best
  possible path through the word-graph, based on acoustic scores only
  (Possible); a combination of acoustic score and bigram score 
  (Acoustic + Bigram), as reported by the current version of the
  system.}
\end{table*}

\subsection{Concept Accuracy}

Word accuracy provides a measure for the extent to which linguistic
processing contributes to speech recognition. However, since the main task
of the linguistic component is to analyze utterances semantically, an
equally important measure is {\em concept accuracy}, i.e.\ the extent to
which semantic analysis corresponds with the meaning of the 
utterance that was actually produced by the user.

For determining concept accuracy, we have used a semantically
annotated corpus of 3800 user responses. 
Each user response was annotated with an {\em update} representing  
the meaning of the utterance that was actually spoken. 
The annotations were
made by our project partners in Amsterdam, in accordance with the
guidelines given in \cite{VE96}.

Updates take the form described in Section~\ref{dialogue}. 
An update is a logical
formula which can be evaluated against an information state and which 
gives rise to a new, updated information state. The most
straightforward method for evaluating concept accuracy in this setting
is to compare (the normal form of) the update produced by the grammar
with (the normal form of) the annotated update. A major obstacle for 
this approach, however, is the fact that very fine-grained semantic 
distinctions can be made in the update-language. While these
distinctions are relevant semantically (i.e. in certain cases they
may lead to slightly different updates of an information state), they
often can be ignored by a dialogue manager. For instance, the update
below is semantically not equivalent to the one given in 
Section~\ref{dialogue}, as
the ground-focus distinction is slightly different. 

\begin{verbatim}
userwants.travel.destination.place
                  ([# town.leiden];
                   [! town.abcoude])
\end{verbatim}

\noindent However, the dialogue manager will decide in both cases that
this is a correction of the destination town. 

Since semantic analysis is the input for the dialogue manager, we have
therefore measured concept accuracy in terms of a simplified version
of the update language. Following the proposal in \cite{BO96}, we
translate each update into a set of {\em semantic units}, were a unit
in our case is a triple $\langle${\tt CommunicativeFunction, Slot,
  Value}$\rangle$. For instance, the example above, as well as the
example in Section~\ref{dialogue}, translates as 

\noindent
$\langle$ {\tt denial, destination\_town, leiden} $\rangle$\\
$\langle$ {\tt correction, destination\_town, abcoude} $\rangle$

\noindent
Both the updates in the annotated corpus and the updates produced by
the system were translated into semantic units of the form given
above. 

Semantic accuracy is given in the following tables according to four
different definitions. Firstly, we list the proportion of utterances
for which the corresponding semantic units exactly match the semantic
units of the annotation ({\em match}). Furthermore we calculate 
{\em precision} (the number of correct semantic units divided by the
number of semantic units which were produced) and {\em recall} 
(the number of correct semantic units divided by the number of semantic
units of the annotation). 
Finally, following \cite{BO96}, we also present concept accuracy as

\[
CA = 100 \left( 1 - \frac{SU_S + SU_I + SU_D}{SU} \right) \%
\] 

\noindent 
where $SU$ is the total number of semantic units in the translated corpus
annotation, and $SU_S$, $SU_I$, and $SU_D$ are the number of
substitutions, insertions, and deletions that are necessary to make
the translated grammar update equivalent to the translation of the 
corpus update. 

We obtained the results given in Table~\ref{tabc}.

\begin{table*}[tb]
\begin{center}
\begin{tabular}{|l|c|r|r|r|r|r|r|}\hline
& Method         &    WA &    SA & \multicolumn{4}{|c|}{Semantic accuracy}\\
\cline{5-8}  
&               &       &       & match & precision & recall & CA  \\\hline
{\em 3800 graphs:} &
user utterance &       &       &  97.9 &     99.2  &   98.5 & 98.5\\
& word-graphs    &  85.3 &  72.9 &  81.0 &     84.7  &   86.6 & 84.4\\
& word-graphs (+bigram)
               &  86.5 &  75.1 &  81.8 &     85.5  &   87.4 & 85.2\\\hline
{\em 5000 graphs:} &
word-graphs    &  79.5 &  70.0 &       &           &        &\\
& word-graphs (+bigram)
               &  82.4 &  74.2 &       &           &        &\\\hline
\end{tabular}
\end{center}
\caption{\label{tabc}
  Evaluation of the NLP component with respect to word accuracy,
  sentence accuracy and concept accuracy. Semantic accuracy consists
  of the percentage of graphs which receive a fully correct analysis
  (match), percentages for precision and recall of semantic slots, and
  concept accuracy.  The first row presents the results if the parser
  is given the actual user utterance (obviously WA and SA are
  meaningless in this case).  The second and third rows present the
  results for word-graphs. In the third row bigram information is
  incorporated in the robustness component.}
\end{table*}

The following reservations should be made with respect to the numbers given
above.

\begin{itemize}
\item The test set is not fully representative of the task, because
  the word-graphs are relatively simple.
\item The test set was also used during the design of the
  grammar. Therefore the experiment is methodologically unsound since
  no clear separation exists between training and test material.
\item Errors in the annotated corpus were corrected by us.
\item Irrelevant differences between annotation and analysis were
  ignored (for example in the case of the station names {\em cuijk}
  and {\em cuyk}). 
\end{itemize}

Even if we take into account these reservations, it seems that we
can conclude that the robustness component adequately extracts useful
information even in cases where no full parse is possible: concept
accuracy is (luckily) much higher than sentence accuracy.

\section*{Conclusion}

We have argued in this paper that sophisticated grammatical analysis
in combination with a robust parser can be applied successfully as an
ingredient of a spoken dialogue system.  Grammatical analysis
is thereby shown to be a viable alternative to techniques such as concept
spotting. We showed that for a state-of-the-art application (public
transport information system) grammatical analysis can be applied
efficiently and effectively. It is expected that the use of
sophisticated grammatical analysis allows for easier construction of
linguistically more complex spoken dialogue systems.

\section*{Acknowledgments}

This research is being carried out
within the framework of the Priority Programme
Language and Speech Technology (TST). The TST-Programme is
sponsored by NWO (Dutch Organization for Scientific Research).


\end{document}